\documentclass{article}

\usepackage[preprint]{neurips_2024}

\usepackage[utf8]{inputenc} 
\usepackage[T1]{fontenc}    
\usepackage{hyperref}       
\usepackage{url}            
\usepackage{booktabs}       
\usepackage{amsfonts}       
\usepackage{nicefrac}       
\usepackage{microtype}      
\usepackage{xcolor}         

\usepackage{tabularx}
\usepackage{multirow}
\usepackage{booktabs}
\usepackage{tikz}
\usepackage{hyperref}
\usepackage{amsmath}
\usepackage{CJKutf8}
\usepackage{longtable}
\usepackage{ltxtable}
\usepackage{subcaption}
\usepackage{caption}
\usepackage{makecell}
\usepackage{xltabular}
\usepackage{array}
\usepackage{longtable}
\usepackage{tabu}
\usepackage{makecell}
\usepackage{lipsum}

\title{StepMathAgent: A Step-Wise Agent for Evaluating Mathematical Processes through Tree-of-Error}

\author{
Shu-Xun Yang$^{1\dagger}$ , Cunxiang Wang$^{1\dagger}$, Yidong Wang$^{1}$, Xiaotao Gu$^{1}$, Minlie Huang$^{2}$, Jie Tang$^{2}$
\\ \\
$^1$Zhipu.AI \quad $^2$Tsinghua University \\
}

\begin{document}

\maketitle

\begin{abstract}
Evaluating mathematical capabilities is critical for assessing the overall performance of large language models (LLMs). However, existing evaluation methods often focus solely on final answers, resulting in highly inaccurate and uninterpretable evaluation outcomes, as well as their failure to assess proof or open-ended problems. To address these issues, we propose a novel mathematical process evaluation agent based on Tree-of-Error, called \textbf{StepMathAgent}. This agent incorporates four internal core operations: logical step segmentation, step scoring, score aggregation and error tree generation, along with four external extension modules: difficulty calibration, simplicity evaluation, completeness validation and format assessment. Furthermore, we introduce \textbf{StepMathBench}, a benchmark comprising 1,000 step-divided process evaluation instances, derived from 200 high-quality math problems grouped by problem type, subject category and difficulty level. Experiments on StepMathBench show that our proposed StepMathAgent outperforms all state-of-the-art methods, demonstrating human-aligned evaluation preferences and broad applicability to various scenarios. Our data and code are available at https://github.com/SHU-XUN/StepMathAgent. 
\end{abstract}

\thispagestyle{plain}
\renewcommand{\thefootnote}{\fnsymbol{footnote}}
    \footnotetext[2]{Contributed equally.}
\renewcommand{\thefootnote}{\arabic{footnote}}

\section{Introduction}

Large language models (LLMs) have recently made remarkable progress, driving advancements across a wide range of industries and applications \citep{survey-llm}. Evaluating the mathematical capabilities of LLMs is a critical aspect of assessing their overall performance \citep{survey-evaluation}. Currently, numerous evaluation efforts \citep{evaluation1,evaluation2,evaluation3,evaluation4} have been conducted on various mathematical datasets \citep{dataste1,dataset2,dataset3,dataset4} constructed across different dimensions.

\begin{figure}[htb]
  \centering
  \includegraphics[width=\linewidth]{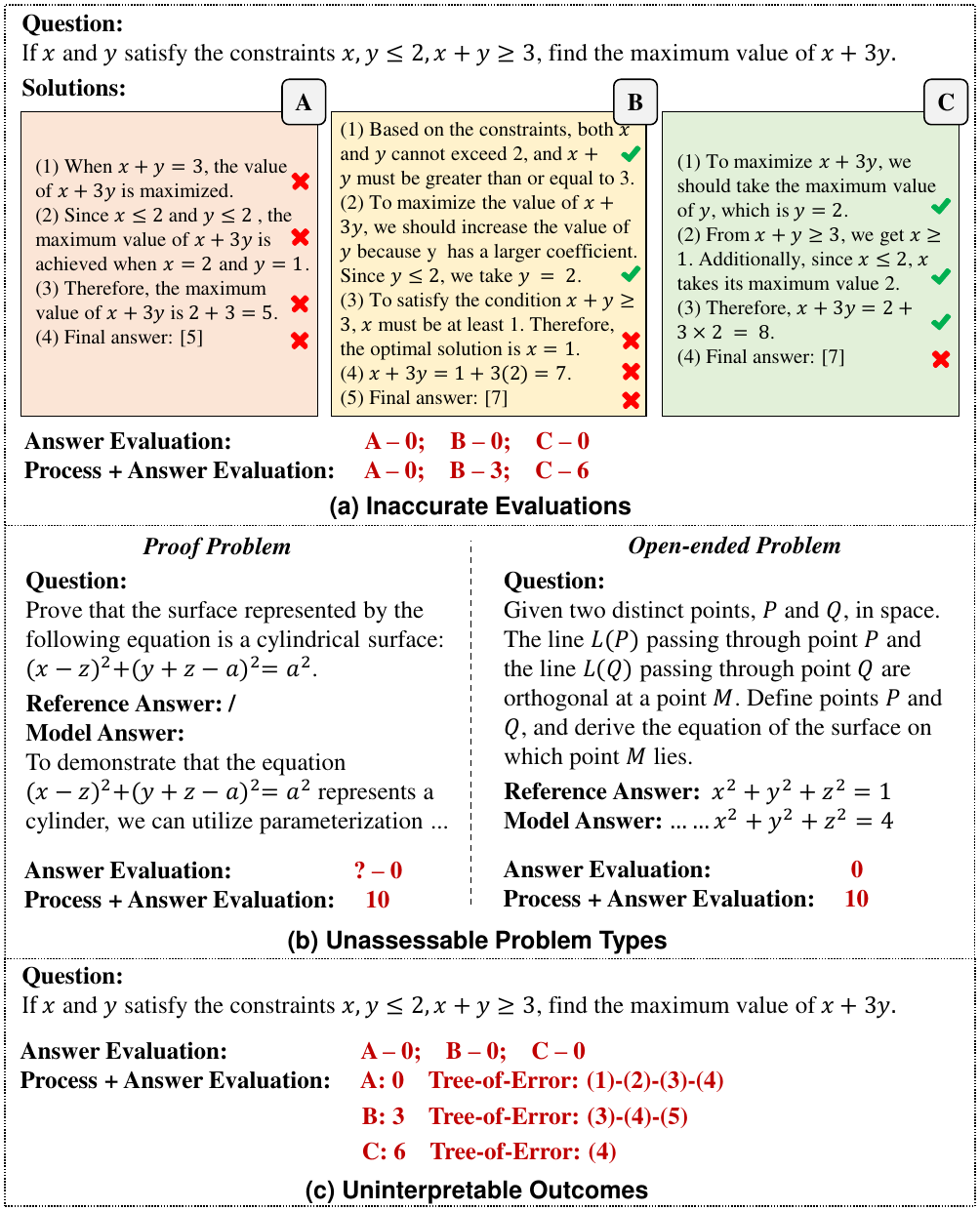}
  \caption{Critical issues arising from answer evaluation.}
  \label{fig1}
\end{figure}

However, existing evaluation methods {\citep{evaluation1,evaluation2,evaluation3,evaluation4}} predominantly focus on the final problem-solving answers, while neglecting the fine-grained assessments of the problem-solving process \citep{GSM8K}. As shown in Figure \ref{fig1}, this limitation gives rise to several critical issues: (1) \textbf{Inaccurate Evaluations}. In fact, relying solely on answers for evaluation often fails to accurately reflecting the true capabilities of LLMs. For instance, it is highly unreasonable for a model with both an incorrect problem-solving process and final answer to receive the same score as a model with a correct process but an incorrect answer.
(2) \textbf{Unassessable Problem Types}. Obviously, current evaluation methods depending on final answers are inadequate for assessing mathematical proof problems that emphasize the logic and rigor of the problem-solving process, or open-ended problems that do not have a unique correct answer.
(3) \textbf{Uninterpretable Outcomes}. If evaluation relies solely on final answers, it becomes impossible to pinpoint the specific steps at which LLMs make errors during problem-solving or to track their mistakes throughout the evaluation process, leading to low interpretability and limited feedback. In summary, performing fine-grained evaluations of the problem-solving process is of paramount importance.

In the practical application of process evaluations, several challenges must be addressed: (1) \textit{Varied Problem-Solving Methods}. There are numerous approaches to solving the same mathematical problem, and it is impractical to enumerate them all.
(2) \textit{Mixed Process Correctness}. The problem-solving process often involves a mixture of correct and incorrect steps, with limited distinguishability and poor readability.
(3) \textit{Diverse Application Priorities.} Different evaluation scenarios impose varying requirements, where some emphasize conciseness while others prioritize logical completeness. 
(4) \textit{Sparse Datasets}. Process evaluation datasets in the mathematical domain are scarce, as their construction is highly challenging and costly.

In this paper, we propose \textbf{StepMathAgent}, a general and novel mathematical process evaluation agent that resolves \textit{Challenge (1)} without relying on any scoring criteria or reference answers. This agent builds upon {Tree-of-Error} and incorporates four internal core operations, i.e., logical step segmentation, step scoring, score aggregation and error tree generation, effectively tackling \textit{Challenge (2)}. Meanwhile, four external extension modules are integrated in this agent, i.e., difficulty calibration, simplicity evaluation, completeness validation and format assessment, to overcome \textit{Challenge (3)}.
Furthermore, we introduce a mathematical process evaluation benchmark for \textit{Challenge (4)}, called StepMathBench. This benchmark consists of 1,000 step-divided process evaluation instances and 200 high-quality math problems, which is divided by problem type, subject category and difficulty level. Experiments on StepMathBench prove that StepMathAgent outperforms other baselines and achieves the state-of-the-art performance. 
Moreover, the error trees generated by StepMathAgent not only enhance interpretability and provide extensive feedback but also offer profound insights into the optimization of LLMs' reasoning capabilities. By assisting in identifying cases such as minor deviations in the early stages of reasoning propagating into significant overarching errors, LLMs deriving correct answers through guessing despite the absence of logical reasoning, and fully accurate reasoning processes still resulting in subtle errors in the final answer, the error trees facilitate a deeper understanding of the inherent limitations and behaviors of LLMs. These findings demonstrate that StepMathAgent, while achieving a high degree of alignment with human scoring preferences, is well-suited for various process evaluation scenarios.

In summary, our contributions are as follows:
\begin{itemize}
\item We are the first to systematically analyze the limitations and challenges of mathematical evaluation and to offer a novel perspective on general mathematical process evaluation.
\item We present the novel StepMathAgent based on {Tree-of-Error} for process evaluation, where four internal core operations and four external extension modules are fully integrated.
\item We construct StepMathBench for mathematical process evaluation. Experiments demonstrate that StepMathAgent performs significantly better than state-of-the-art baselines.
\end{itemize}

\section{Related Work}

\subsection{Mathematical Evaluation}

In recent years, the field of mathematical evaluation has witnessed significant growth. 
Numerous datasets are constructed in this domain, such as GSM8K \citep{GSM8K}, MATH \citep{MATH}, MATH-140 \citep{MATH-140}, TabMWP \citep{TabMWP} and MathQA \citep{MathQA}, establishing a solid foundation for further research. Building on these efforts, several innovative approaches to evaluation are introduced. For example, \citet{evaluation1} introduce a robust evaluation framework designed to assess large language models (LLMs) through extensive unit tests, with a focus on both the accuracy and generality of model responses. 
\citet{evaluation3} design a checklist for testing task generalization and reasoning robustness to judge if a model really understands a problem, as well as an automatic tool to generate checklists efficiently.
\citet{evaluation4} propose a comprehensive mathematical
evaluation toolkit that not only utilizes a python computer algebra system for its numerical accuracy, but also integrates an optional LLM for evaluation. 
Despite the effectiveness of these methods, the mainstream approaches \citep{survey-math} for evaluating the outputs of LLMs remain limited to rule-based exact matching and LLM-based scoring, both of which predominantly rely on final answers.
While LLM-based scoring can be guided by prompts to focus on the process, this coarse-grained strategy fails to address numerous challenges discussed above.
Different from these methods, in this paper, we propose a step-wise benchmark and a novel general-purpose agent for mathematical process evaluation, capable of performing fine-grained assessments by integrating both problem-solving processes and final answers. This approach ensures precise mathematical process evaluation across diverse problem types, while also delivering full interpretability and comprehensive feedback, thereby making it highly adaptable to a wide range of real-world scenarios.

\begin{table}[tb]
  \small
  \begin{tabularx}{\linewidth}{X|l|X|l}
    \toprule
    \textbf{Primary Category} & \textbf{Secondary Category} & \textbf{Primary Category} & \textbf{Secondary Category} \\
    \midrule
    \multirow{3}{*}[-0ex]{\parbox{2cm}{Elementary \\ Mathematics}} & Arithmetic & \multirow{3}{*}[-0ex]{Modern Mathematics} & Advanced Mathematics\\
    & Algebra & & Linear Algebra\\
    & Geometry & & Analytic Geometry \\
    \midrule
    \multirow{4}{*}[-0ex]{\parbox{2cm}{Contemporary \\ Mathematics}} & Discrete Mathematics & \multirow{5}{*}[-0ex]{Applied Mathematics} &
    Financial and Economic \\
    & Probability and Statistics & & Real-World Applications \\
    &  Number Theory & & Optimization and Planning \\
    & Functional Analysis & & Misguidance \\
    & & & Other Applications \\
    
    \bottomrule
\end{tabularx}
\vspace{0.3cm}
\caption{Subject categories in the field of mathematics.}
\label{tab1: category}
\end{table}

\begin{table}[tb]
  \small
  \centering
  \begin{tabularx}{.6\linewidth}{l|X|X|X|X|X}
    \toprule
    \multicolumn{3}{c|}{\textbf{Knowledge}} & \multicolumn{3}{c}{\textbf{Solution Steps}} \\
    \midrule
    \textbf{Scope} & \textbf{Point} & \textbf{Level} & \textbf{1-2} & \textbf{3-5} & \textbf{>5} \\
    \midrule
    \multirow{3}{*}[-0ex]{$\leq$ Middle School} & 1 & 1 & \underline{1} & \underline{1} & \underline{2} \\
    & 2 & 1 & \underline{1} & \underline{1} & \underline{2} \\
    & 3 & 2 & \underline{1} & \underline{2} & \underline{3} \\
    \midrule
    \multirow{3}{*}[-0ex]{$=$ High School} & 1 & 2 & \underline{1} & \underline{2} & \underline{3} \\
    & 2 & 2 & \underline{1} & \underline{2} & \underline{3} \\
    & 3 & 3 & \underline{2} & \underline{3} & \underline{3} \\
    \midrule
    \multirow{3}{*}[-0ex]{$\geq$ Undergraduate} & 1 & 2 & \underline{1} & \underline{2} & \underline{3} \\
    & 2 & 3 & \underline{2} & \underline{3} & \underline{3} \\
    & 3 & 3 & \underline{2} & \underline{3} & \underline{3} \\
    \bottomrule
\end{tabularx}
\vspace{0.3cm}
\caption{Difficulty classification system for mathematical problems based on knowledge and solution, where 1 denotes easy, 2 denotes medium and 3 denotes hard.}
\label{tab2: level}
\end{table}

\subsection{Process-based Research}

In fact, the process plays a pivotal role in both evaluation and reasoning, garnering significant attention from researchers \citep{process1,process2}. Typically, process-level reward models (PRMs) are widely employed to enhance the process-level reasoning accuracy of LLMs \citep{process3,process4,process5}. For instance, \citet{process6} propose a coarse-to-fine process reward modeling framework to enhance mathematical reasoning. \citet{process7} design a retrieval-augmented reasoning through trustworthy process rewarding framework that enhances reasoning capabilities via post-training and test-time scaling. \citet{process8} introduce a challenging process-level benchmark specifically designed to assess the fine-grained error detection capabilities of PRMs. However, to the best of our knowledge, there is currently little to no work that effectively and fully integrates the process into mathematical evaluation frameworks. ProcessBench \citep{process9} attempt to measure the ability of LLMs to identify erroneous steps in mathematical reasoning, yet their scope is restricted to detecting the first erroneous step, rather than establishing a general and comprehensive framework for process evaluation. In this paper, we systematically analyze the limitations and challenges of mathematical evaluation for the first time and offer a novel perspective on general mathematical process evaluation, centered around StepMathAgent and StepMathBench.



\section{Dataset}

\begin{figure}[tb]
  \centering
  \includegraphics[width=0.6\linewidth]{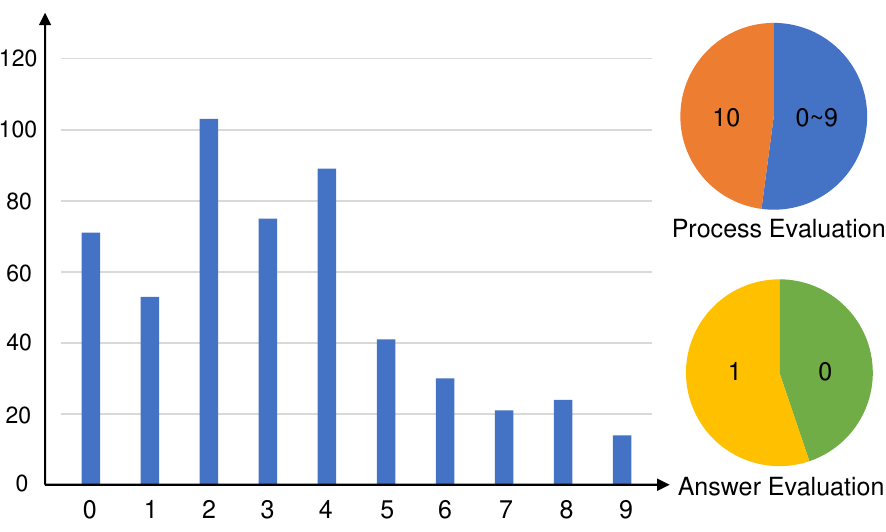}
  \caption{Distribution of solution scores.}
  \label{fig2}
\end{figure}

Given a mathematical problem $P$ and a solution $S$ encompassing both the problem-solving process and the final answer, the objective of mathematical process evaluation is to provide a comprehensive assessment of the solution by simultaneously considering its process and answer, and to output a score $G \in [0,10]$. Notably, the reference answer $R$ is treated as an optional input.

Based on this task definition, we first construct 200 high-quality and unique mathematical problems divided by problem type, subject category and difficulty level for StepMathBench, ensuring extensive coverage of various types while eliminating the risk of data leakage. 
Problem types include calculation, proof and open-ended problems, with a proportional distribution of 29:10:1. Subject categories span all educational levels, from elementary, middle and high school to undergraduate and higher. Further details are provided in Table \ref{tab1: category}. When calculated according to the primary category, the ratio of the four subject categories is 53:40:67:40. Difficulty levels are classified into easy, medium and hard, taking into account the knowledge scope, the complexity of knowledge points and the number of coarse-grained standardized solution steps. Detailed classification criteria can be found in Table \ref{tab2: level}. The three difficulty levels are distributed in a ratio of 4:15:31. Besides, each problem is assigned a constraint attribute that restricts the output of large language models (LLMs), enabling rule-based exact matching evaluations.

Then, we generate solutions for each mathematical problem using widely adopted LLMs, including GPT-4o (gpt-4o-2024-08-06) \citep{gpt4}, o1-mini (o1-mini-2024-09-12) \citep{gpt4}, Claude-3.5 (claude-3.5-sonnet-2024-06-20) \citep{claude}, Gemini-1.5 (gemini-1.5-pro2-002) \citep{gemini} and Llama-3.1 (llama-3.1-70b-instruct-turbo) \citep{llama}, ultimately producing 1,000 distinct solutions in StepMathBench. 
For calculation problems, solutions generated by LLMs are required to include both the detailed problem-solving process and the final answer, whereas no specific requirements are imposed on the solutions for proof and open-ended problems. 
Subsequently, we engage six annotators, all holding at least a master’s degree, to segment and score each solution. The segmentation principle is to divide the reasoning solution into the smallest possible fine-grained steps while ensuring the retention of meaningful information. 
The scoring system includes three labels: correct, incorrect and correct-but-meaningless. 
The correct-but-meaningless label applies to reasoning steps that are correct in isolation but derive from previously incorrect reasoning steps, rendering them meaningless from the perspective of the overall solution. This label is specifically introduced to better facilitate the construction of error trees.

Finally, we simulate human scoring preferences to compute the final score for each solution $\bar{S} = \{s_1, s_2, ..., s_N\}$ of length $N$ with step scores $\bar{S}_G = \{g_1, g_2, ..., g_N\}$.
Steps labeled as correct are assigned a score of 1, incorrect receive a score of 0, and correct-but-meaningless are also given a score of 0. For calculation problems, the final score $\bar{G}_C$ is determined with a weighted ratio of 6:4 between the process and the final answer, while the score for proof problems $\bar{G}_P$ and open-ended problems $\bar{G}_O$ is computed by applying weights to the process:

\vspace{-0.4cm}
\begin{equation} \label{eq1: score}
\begin{split}
&\bar{G}_C = \phi(6 * \frac{\sum_{i=1}^{N-1} {\bar{G}_i}}{N-1} + 4 * g_N) \\
&\bar{G}_P = \bar{G}_O = \phi(10 * \frac{\sum_{i=1}^{N} {\bar{G}_i}}{N})
\end{split}
\end{equation}
where $\phi$ denotes the rounding operation. In particular, to assess the performance of rule-based exact matching evaluation methods, we additionally assign each solution a score $\bar{G}' \in \{0, 1\}$, determined solely by the final result. The score for calculation problems $\bar{G}'_C$ is equal to $g_N$. If the scores $\bar{G}_P$ and $\bar{G}_O$ are less than or equal to 5, then $\bar{G}'_P = \bar{G}'_C = 0$, otherwise, $\bar{G}'_P = \bar{G}'_C = 1$. In multiple cross-check validations, the inter-annotator agreement achieves a high level of 95\%. The distribution of solution scores is shown in Figure \ref{fig2}, while data samples and prompts are provided in Appendix \ref{sec:appendix dataset}.

\begin{figure*}[t]
  \centering
  \includegraphics[width=\textwidth]{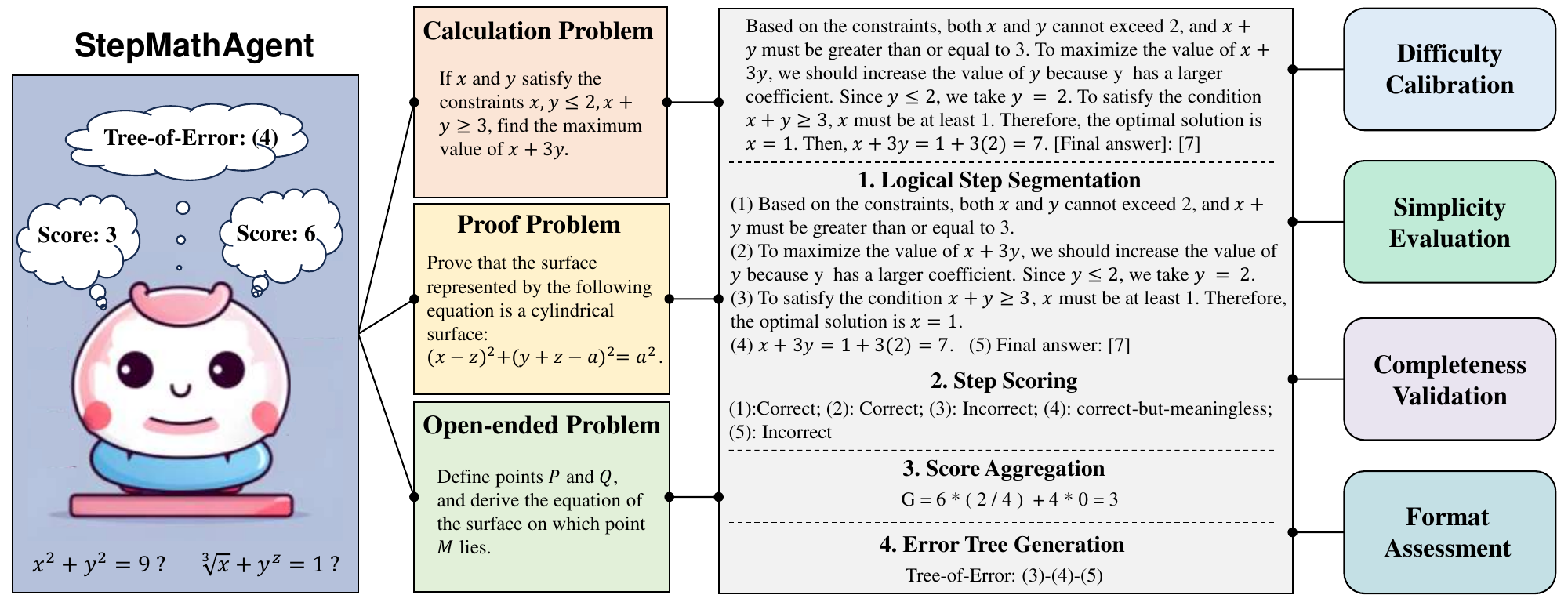}
  \caption{The overall architecture of StepMathAgent.}
  \label{fig3}
  \vspace{0mm}
\end{figure*}

\section{Method}

In this paper, we propose a novel mathematical process evaluation agent based on {Tree-of-Error}, named StepMathAgent. The overall architecture of StepMathAgent is illustrated in Figure \ref{fig3}, which is composed of four internal core operations and four external extension modules.

\subsection{Internal Operations}

The core internal operations of StepMathAgent are fundamentally designed to emulate the human process evaluation workflow, specifically including logical step segmentation, step scoring, score aggregation and error tree generation.

\paragraph{Logical Step Segmentation} Given a mathematical problem $P$ and its solution $S$, the purpose of the logical step segmentation operation is to comprehensively analyze the logical structure of the solution and systematically segment it into a well-defined set of steps $S = \{s_1, s_2, ..., s_N\}$.

\paragraph{Step Scoring} Following this, the step scoring operation performs an objective evaluation of each segmented step and assigns a corresponding score. This task can be formulated as a three-class classification problem, with labels aligned with those defined during dataset construction, i.e., correct, incorrect and correct-but-meaningless. Then, the step scoring operation produces a score set $S_G = \{g_1, g_2, ..., g_N\}$, where each score $g_i$ corresponds directly to a segmented step $s_i$.

\paragraph{Score Aggregation} Subsequently, the score aggregation operation distinguishes among calculation, proof and open-ended problems, and computes the final score $G$ for the solution $S$ based on the segmented steps and their corresponding scores $\{S \rightarrow S_G\}$ using Equation (\ref{eq1: score}).

\paragraph{Error Tree Generation} Finally, the error tree generation operation comprehensively analyzes the segmented steps and their corresponding scores $\{S \rightarrow S_G\}$ to identify all possible error chains $T$ and construct a {Tree-of-Error}, thereby enhancing interpretability and providing detailed feedback.

\subsection{External Modules}

In contrast to the internal operations, the primary function of the external extension modules is to accommodate diverse real-world evaluation scenarios by offering additional functionalities that enhance their applicability. These modules consist of difficulty calibration, simplicity evaluation, completeness validation and format assessment.

\paragraph{Difficulty Calibration} Considering that some extremely simple mathematical problems can be adequately evaluated by solely determining the correctness of the final answer without the need for process evaluation, the difficulty calibration module is specifically designed to determine whether a problem is sufficiently simple to bypass process evaluation. If process evaluation is deemed unnecessary, only the correctness of the answer is verified. Otherwise, it activates the four core internal operations in StepMathAgent to conduct a comprehensive process evaluation. This difficulty calibration module is primarily designed to conserve resources and enhance evaluation efficiency.

\paragraph{Simplicity Evaluation} The simplicity evaluation module is specifically designed for scenarios that prioritize process simplicity. Its operation entails classifying any step as incorrect if it is deemed redundant or devoid of meaningful information.

\paragraph{Completeness Validation} The completeness validation module is specifically tailored for scenarios that prioritize the completeness of reasoning. In this context, when assessing the correctness of segmented steps, any step with an incomplete reasoning chain, such as missing premises, missing conclusions or skipped steps, is deemed incorrect.

\paragraph{Format Assessment} The format assessment module is 
tailored for scenarios with stringent formatting requirements. In such cases, if a segmented reasoning step includes a LaTeX formula with formatting errors, the step is deemed incorrect.

\begin{table*}[tb]
    \centering
    \small
    \begin{tabularx}{\linewidth}{lXXXXXXXXXXXXXXXXX}
         \toprule
        & \multicolumn{4}{c}{\textbf{All}} & \multicolumn{4}{c}{\textbf{Calculation}} & \multicolumn{4}{c}{\textbf{Proof}} & \multicolumn{4}{c}{\textbf{Open-ended}} \\
        \cmidrule(lr){2-5} \cmidrule(lr){6-9} \cmidrule(lr){10-13} \cmidrule(lr){14-17}
        & \textbf{AvgS} & \textbf{Corr} & \textbf{MSE} & \textbf{OR} & \textbf{AvgS} & \textbf{Corr} & \textbf{MSE} & \textbf{OR} & \textbf{AvgS} & \textbf{Corr} & \textbf{MSE} & \textbf{OR} & \textbf{AvgS} & \textbf{Corr} & \textbf{MSE} & \textbf{OR} \\
        \midrule
        \addlinespace[3pt]
        \multicolumn{17}{c}{\textit{\textbf{Final Answer Evaluation}}} \\[3pt]
        RuleEM & - & - & - & - & 39.0 & 72.6 & 14.6 & 85.4 & - & - & - & - & - & - & - & - \\
        GPT-4o$_{V1}$ & 57.8 & 77.7 & 11.0 & 89.0 & 48.6 & 90.4 & 4.8 & 95.2 & 87.2 & 28.5 & 25.6 & 74.4 & 32.0 & 8.3 & 44.0 & 56.0 \\
        o1-mini$_{V1}$ & 56.2 & 77.7 & 11.0 & 89.0 & 50.3 & 92.3 & 3.9 & 96.1 & 76.8 & 25.7 & 28.8 & 71.2 & 20.0 & 16.1 & 40.0 & 60.0\\
        Claude-3.5$_{V1}$ & 62.9 & 75.7 & 12.3 & 87.7 & 52.4 & 88.8 & 5.7 & 94.3 & 95.6 & 7.9 & 29.2 & 70.8 & 40.0 & 26.3 & 36.0 & 64.0 \\
        GLM-4$_{V1}$ & 57.1 & 78.3 & 10.7 & 89.3 & 48.6 & 91.5 & 4.3 & 95.7 & 86.0 & 25.3 & 26.8 & 73.2 & 16.0 & 27.3 & 36.0 & 64.0 \\
        Qwen-turbo$_{V1}$ & 59.9 & 75.6 & 12.1 & 87.9 & 49.9 & 87.0 & 6.5 & 93.5 & 91.6 & 22.1 & 26.8 & 73.2 & 32.0 & 42.8 & 28.0 & 72.0\\
        \midrule
        \textbf{Gold} & 55.2 & 100 & 0 & 100 & 50.1 & 100 & 0 & 100 & 71.2 & 100 & 0 & 100 & 44.0 & 100 & 0 & 100\\
        \midrule
        \addlinespace[3pt]
        \multicolumn{17}{c}{\textit{\textbf{Problem-solving Process Evaluation}}} \\[3pt]
        GPT-4o$_{V2}$ & 73.4 & \underline{70.2} & \underline{8.0} & 36.9 & 71.1 & \underline{75.7} & \underline{7.3} & 39.4 & 81.7 & \textbf{41.3} & 10.1 & 30.8 & \underline{57.2} & \textbf{70.8} & \textbf{6.1} & \underline{24.0}\\
        o1-mini$_{V2}$ & \textbf{65.6} & 61.6 & 10.2 & 41.4 & 65.9 & 74.2 & \underline{7.3} & \underline{47.2} & 66.8 & 23.8 & 18.4 & 27.6 & 43.4 & 47.1 & 11.6 & 12.0\\
        Claude-3.5$_{V2}$ & 82.5 & 51.0 & 13.7 & 20.5 & 80.9 & 55.3 & 14.1 & 23.9 & 87.8 & 24.8 & 12.0 & 12.8 & 75.6 & 25.8 & 17.6 & 0 \\
        GLM-4$_{V2}$ & 72.6 & 67.4 & 8.5 & 16.1 & 70.9 & 72.7 & 8.2 & 19.3 & 78.6 & \underline{36.0} & \textbf{9.4} & 6.8 & 63.6 & 56.0 & \underline{9.1} & 16.0 \\
        Qwen-turbo$_{V2}$ & 76.3 & 63.0 & 10.0 & 23.0 & 74.5 & 67.2 & 9.9 & 27.6 & 82.3 & \underline{36.0} & \underline{10.0} & 10.8 & 69.6 & 59.0 & 10.3 & 12.0\\
        \midrule
        \textbf{StepMathAgent} & \underline{66.2} & 51.4 & 12.4 & 38.9 & \textbf{63.4} & 57.1 & 11.7 & 42.5 & \textbf{75.1} & 21.3 & 14.6 & 30.4 & \textbf{56.8} & \underline{53.3} & 9.7 & 20.0\\
        StepMathAgent$_O$ & 68.2 & \textbf{70.5} & \textbf{7.9} & \textbf{48.1} & \underline{64.7} & \textbf{78.4} & \textbf{6.2} & \textbf{53.1} & 78.6 & 34.9 & 12.3 & \underline{35.6} & 67.2 & 51.2 & 13.0 & \textbf{28.0} \\
        StepMathAgent$_C$ & 83.1 & 50.7 & 14.3 & \underline{44.0} & 79.2 & 54.3 & 14.0 & 44.7 & 94.5 & 22.1 & 14.5 & \textbf{44.8} & 82.8 & 31.1 & 20.7 & 16.0 \\
        StepMathAgent$_G$ & 72.3 & 53.4 & 11.6 & 39.5 & 67.6 & 59.1 & 10.8 & 41.8 & 85.7 & 19.4 & 13.9 & 34.4 & 74.8 & 52.0 & 13.3 & \underline{24.0}\\
        StepMathAgent$_Q$ & 60.2 & 48.2 & 14.7 & 33.1 & 57.2 & 53.1 & 13.8 & 35.2 & \underline{70.2} & 28.1 & 16.6 & 28.8 & 47.2 & 4.8 & 21.1 & 16.0 \\
        \midrule
        \textbf{Gold} & 64.8 & 100 & 0 & 100 & 62.0 & 100 & 0 & 100 & 74.2 & 100 & 0 & 100 & 52.8 & 100 & 0 & 100 \\
         \bottomrule
    \end{tabularx}
    \caption{Results on StepMathBench in terms of AvgS, Corr, MSE and OR. Each metric is normalized to a range of 0-100. LLMs labeled as $_{V1}$ and $_{V2}$ correspond to answer-based evaluation and process-aware evaluation. The core \textbf{StepMathAgent} is implemented with GPT-4o, while StepMathAgent$_O$, StepMathAgent$_C$, StepMathAgent$_G$ and StepMathAgent$_Q$ are implemented using o1-mini, Claude-3.5, GLM-4 and Qwen-turbo, respectively. Besides, \textbf{bold} indicates the best performance, while \underline{underline} represents the second best performance.}
    \label{tab3: overall results}
    \vspace{0mm}
\end{table*}

\section{Experiment}
\subsection{Settings}
\paragraph{\textbf{Dataset and Evaluation Metrics}}

We evaluate the performance of our proposed StepMathAgent and eleven baselines on the constructed StepMathBench. Average score (AvgS), correlation coefficient (Corr), mean squared error (MSE) and overlap rate (OR) are adopted as evaluation metrics to comprehensively assess their performance. 
Specifically, AvgS represents the average score $(\sum_{i=1}^{i=K} G_i)/K$ assigned by evaluation model across the entire StepMathBench, where a value closer to the gold score $(\sum_{i=1}^{i=K} \bar{G}_i)/K$ indicates a better-performing evaluation model. Corr denotes the correlation between the assigned scores $G$ and the gold scores $\bar{G}$, with the value approaching 1 reflecting stronger agreement. MSE quantifies the evaluation error between $G$ and $\bar{G}$, while OR measures the proportion of cases where the assigned scores $G_i$ align exactly with the gold scores $\bar{G}_i$.

\paragraph{\textbf{Baselines}}

We compare our StepMathAgent with eleven baselines covering both rule-based exact matching (RuleEM) methods and LLM-based scoring methods. LLM-based scoring methods is further divided into answer-based LLM evaluation ($V1$), process-aware LLM evaluation ($V2$) and multi-dimensional LLM evaluation ($V3$). Each approach employs five LLMs for assessment, namely, GPT-4o \citep{gpt4}, o1-mini \citep{gpt4}, Claude-3.5 \citep{claude}, GLM-4 \citep{glm} and Qwen \citep{qwen}.

\paragraph{\textbf{Implementation Details}}


Our proposed StepMathAgent can be fully implemented using LLMs, regardless of internal operations or external modules. This agent operates entirely in a zero-shot manner. Consistent with the baseline, we implement our StepMathAgent using five LLMs: GPT-4o (gpt-4o-2024-08-06), o1-mini (o1-mini-2024-09-12), Claude-3.5 (claude-3.5-sonnet-2024-06-20), GLM-4 (glm-4-plus) and Qwen-turbo (qwen-turbo). Besides, in the computation of evaluation metrics, $K = 1000$. All the prompts used in our experiments are provided in Appendix \ref{sec:appendix dataset}.


\subsection{Overall Results}

\begin{table*}[tb]
    \centering
    \small
    \begin{tabularx}{\linewidth}{lXXXXXXXXXXXXXXXXX}
         \toprule
        & \multicolumn{4}{c}{\textbf{All}} & \multicolumn{4}{c}{\textbf{Calculation}} & \multicolumn{4}{c}{\textbf{Proof}} & \multicolumn{4}{c}{\textbf{Open-ended}} \\
        \cmidrule(lr){2-5} \cmidrule(lr){6-9} \cmidrule(lr){10-13} \cmidrule(lr){14-17}
        & \textbf{AvgS} & \textbf{Corr} & \textbf{MSE} & \textbf{OR} & \textbf{AvgS} & \textbf{Corr} & \textbf{MSE} & \textbf{OR} & \textbf{AvgS} & \textbf{Corr} & \textbf{MSE} & \textbf{OR} & \textbf{AvgS} & \textbf{Corr} & \textbf{MSE} & \textbf{OR} \\
        \midrule
         GPT-4o$_{V3}$ & 68.8 & \textbf{73.7} & \textbf{6.7} & 31.6 & 65.8 & \textbf{79.9} & \textbf{5.7} & 35.4 & 78.6 & \textbf{43.6} & \underline{9.7} & 22.4 & \underline{55.8} & \textbf{68.0} & \textbf{6.7} & 12.0\\
        o1-mini$_{V3}$ & 70.7 & \underline{67.5} & 8.4 & \underline{41.1} & 68.2 & 72.8 & 7.7 & \textbf{44.1} & 79.2 & 39.1 & 10.5 & \underline{33.6} & 60.4 & 59.6 & 8.0 & \textbf{28.0} \\
        Claude-3.5$_{V3}$ & 78.8 & 61.8 & 10.8 & 8.4 & 76.2 & 67.7 & 10.3 & 9.8 & 87.3 & 23.4 & 11.9 & 4.4 & 69.7 & 44.0 & 13.2 & 8.0 \\
        GLM-4$_{V3}$ & 69.8 & \textbf{73.7} & \underline{6.8} & 12.8 & 66.6 & \underline{79.3} & \underline{6.0} & 14.2 & 80.0 & \underline{41.1} & \textbf{9.2} & 8.8 & 59.0 & \underline{67.0} & \underline{6.8} & 12.0 \\
        Qwen-turbo$_{V3}$ & 73.2 & 66.9 & 8.6 & 13.0 & 70.5 & 71.3 & 8.3 & 14.5 & 81.6 & 39.6 & 9.5 & 8.8 & 66.6 & 47.5 & 10.9 & 12.0 \\
        \midrule
        
        \textbf{StepMathAgent} & {66.2} & 51.4 & 12.4 & 38.9 & {63.4} & 57.1 & 11.7 & \underline{42.5} & \textbf{75.1} & 21.3 & 14.6 & 30.4 & {56.8} & {53.3} & 9.7 & \underline{20.0}\\
        + Difficulty & 67.2 & 54.4 & 11.8 & \textbf{41.4} & 64.9 & 60.0 & 11.1 & \textbf{44.1} & \underline{75.3} & 24.0 & 14.3 & \textbf{35.6} & \textbf{53.2} & 66.2 & 7.7 & \underline{20.0}\\
        + Simplicity & \underline{65.5} & 51.5 & 12.4 & 37.0 & 63.5 & 58.1 & 11.4 & 40.7 & 72.1 & 24.4 & 14.6 & 28.0 & 56.8 & 16.5 & 18.2 & \underline{20.0}\\
        + Completeness & 62.2 & 50.0 & 13.2 & 37.2 & \textbf{61.3} & 55.5 & 12.4 & 41.4 & 64.9 & 29.0 & 15.3 & 27.2 & 58.0 & 37.0 & 13.5 & 16.0\\
        + Format & \textbf{64.8} & 46.4 & 13.5 & 35.8 & 63.5 & 54.1 & 12.4 & 40.6 & 69.5 & 14.0 & 16.5 & 24.4 & 56.4 & 25.2 & 14.1 & 12.0\\
        + Sim,Com,For & 62.4 & 49.5 & 12.9 & 33.9 & 60.4 & 55.9 & 12.0 & 38.3 & 68.5 & 21.7 & 15.3 & 23.2 & 61.6 & 30.8 & 13.4 & 12.0 \\
        + All & 63.8 & 54.0 & 12.4 & 39.2 & \underline{62.8} & 59.5 & 11.5 & 42.3 & 69.1 & 29.2 & 14.8 & 32.0 & 40.4 & 53.9 & 13.4 & \underline{20.0}\\

        \midrule
        \textbf{Gold} & 64.8 & 100 & 0 & 100 & 62.0 & 100 & 0 & 100 & 74.2 & 100 & 0 & 100 & 52.8 & 100 & 0 & 100 \\
         \bottomrule
    \end{tabularx}
    \caption{Ablation study of external modules in StepMathAgent. LLMs labeled as $_{V3}$ correspond to multi-dimensional evaluation. + indicates the integration of the corresponding external module into the initial StepMathAgent, which is originally composed of only the four internal  core operations. + All signifies the inclusion of all external modules.}
    \label{tab4: ablation study}
\end{table*}

    

Table \ref{tab3: overall results} presents the results of our StepMathAgent and eleven baselines on StepMathBench. From Table \ref{tab3: overall results}, it is evident that final answer evaluation generally yields lower scores than problem-soving process evaluation and fail to accurately capture the true capabilities of LLMs. 
Furthermore, result-based methods, particularly those relying on rule-based exact matching, demonstrate significant limitations in evaluating proof and open-ended problems. 
Even when using LLMs for final answer evaluation, the problem-solving process is often referenced to effectively assess these two types of problems. These observations highlight the critical importance of mathematical process evaluation.

From the perspective of process evaluation, our proposed StepMathAgent exhibits significant advantages. In terms of the evaluation metric AvgS, which quantifies alignment with human scoring preferences, \textbf{StepMathAgent} impletemented by GPT-4o consistently outperforms other models across calculation, proof and open-ended problems, with minimal deviations of only 1.4, 0.9 and 4.0 points from the gold scores, respectively. 
Notably, while the o1-mini$_{V2}$ model achieves an overall AvgS score closer to the gold score, its subcategory scores and other metrics fall significantly short of the performance of StepMathAgent. This suggests that this AvgS score may result from coincidence rather than systematic reliability, thereby lacking strong interpretive validity. Moreover, in Corr, MSE and OR metric, StepMathAgent$_{O}$ substantially surpasses other baselines, showing a score distribution that closely aligns with the gold. These findings highlight the clear superiority of our proposed StepMathAgent in process evaluation.

In fact, we implement StepMathAgent using five distinct LLMs, as detailed in Table \ref{tab3: overall results}, each exhibiting unique characteristics. For example, \textbf{StepMathAgent} and StepMathAgent$_O$ achieves the best overall performance, excelling across all evaluation metrics. StepMathAgent$_Q$ produces scores that are more closely aligned with human preferences but tends to assign lower overall scores.  StepMathAgent$_C$ demonstrates higher OR scores, indicating a greater number of problems where its scores match the gold scores, although its overall scores are higher than expected. StepMathAgent$_G$ achieves higher Corr and lower MSE but also exhibits a tendency to assign inflated scores. Collectively, these results highlight the robustness of our agent. Moreover, regardless of the underlying LLMs employed, these methods with StepMathAgent versions consistently exhibit stronger alignment with human scoring preferences compared to these with ${V2}$ versions, further substantiating the superiority of our proposed agent.

Additionally, we note several intriguing findings from Table \ref{tab3: overall results}. For instance, nearly all models tend to assign inflated scores, with the exception of StepMathAgent$_Q$. This indicates that LLM-based scoring is generally lenient and lacks sufficient strictness. Moreover, as discussed earlier, different models exhibit distinct strengths and characteristics. From the perspective of final answer evaluation, GPT-4o and GLM-4 deliver the strongest overall performance, Qwen-turbo performs well on calculation problems, o1-mini excels on proof problems, and Claude-3.5 is more effective for open-ended problems. In contrast, from the perspective of process evaluation using our proposed StepMathAgent, GPT-4o and o1-mini continue to show exceptional performance overall, while GLM-4 is particularly well-suited for calculation problems, and Qwen-turbo excels in evaluating proof and open-ended problems. Nonetheless, Claude-3.5 proves less suitable as a process evaluation model due to its tendency to assign excessively high scores.

\section{Analysis}
\subsection{Ablation Study of External Modules}

The ablation study results for the four external modules in StepMathAgent are presented in Table \ref{tab4: ablation study}. Each external module effectively demonstrates its intended functionality. For instance, the difficulty calibration module is designed to optimize resource usage and improve evaluation efficiency. In our experiments, approximately 2.3\% of the dataset is identified as overly simple and do not require process evaluation. While the module achieves high precision in classification, its recall is limited due to the fact that around 8\% of the dataset consists of simple problems. Interestingly, we find that incorporating the difficulty calibration module significantly enhances the performance of StepMathAgent on the OR metric. This improvement results from the module's ability to make judgments for simple problems that closely align with human scoring preferences, typically assigning scores of either 0 or 10. As such, this module shows great potential for practical applications.
Additionally, the inclusion of the simplicity evaluation, completeness validation and format assessment modules makes StepMathAgent more stringent during the scoring process, resulting in a reduction in overall scores. Notably, the simplicity evaluation module causes the most 3 points reduction for proof problems, which indirectly reflects the commonly observed phenomenon that proof-solving processes often contain excessive redundant or irrelevant information. The addition of the completeness validation module reduces the scores of calculation problems, bringing them 0.7 points closer to gold scores, while the format assessment module results in an overall score equal to the gold scores, which strongly validate the effectiveness of each module.

Nevertheless, we also observe that incorporating multiple modules simultaneously does not always lead to improved performance. In certain cases, overly stringent evaluations introduce uniformly lower scores. Therefore, in practical applications, it is essential to carefully select modules based on specific requirements and scenarios.
When compared to LLMs with ${V3}$ versions, it can be found that directly employing LLMs for multi-dimensional evaluation yields higher Corr and lower MSE. This approach is suitable for scenarios where a score distribution closely resembling human behavior is desired. Yet, if the objective is to achieve scores that keep a high degree of alignment with human scoring preferences, our StepMathAgent remains the superior choice.




\subsection{Analysis of Step Lengths}


\begin{figure}[tb]
  \centering
  \includegraphics[width=.6\linewidth]{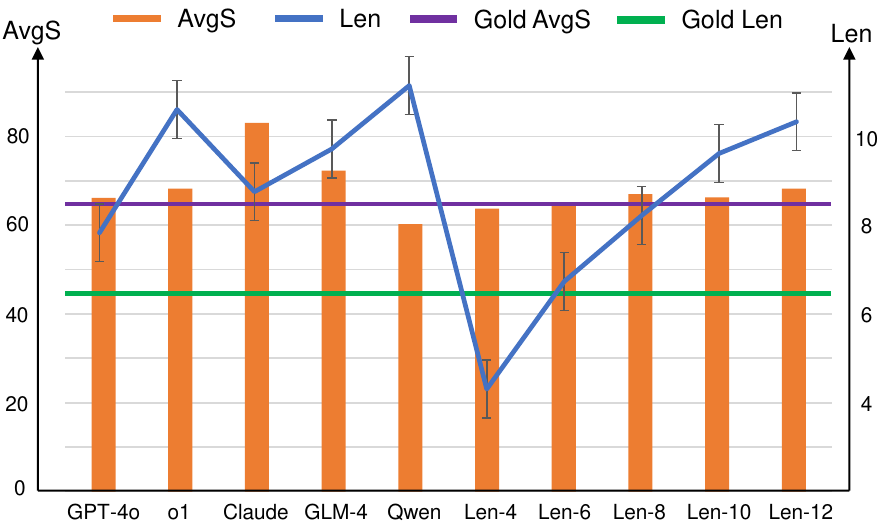}
  \caption{Analysis of AvgS and step lengths.}
  \label{fig4}
\end{figure}

To investigate the relationship between segmented step lengths and the performance of evaluation models, we conduct an analysis of the step lengths produced by StepMathAgent. As illustrated in Figure \ref{fig4}, the steps segmented by StepMathAgent implemented with five different LLMs tend to be longer than those segmented by humans. Nonetheless, whether a negative correlation exists between step length and model performance remains uncertain. To further explore this, we incorporate prior knowledge and use StepMathAgent implemented with GPT-4o to segment problems into approximately 4, 6, 8, 10 and 12 steps, respectively. The results reveal that as the number of segmented steps increases, model scores initially converge toward human scores but eventually diverge as step lengths become excessively long. This finding aligns with the understanding that overly fine-grained segmentations increase the scoring difficulty for LLMs.



\subsection{Case Study}

We analyze several representative examples from StepMathBench, as detailed in Appendix \ref{sec: apppendix case study}. These examples strongly underscore the efficiency and interpretability of the proposed StepMathAgent.

\section{Conclusion}

In this paper, we propose a novel step-wise agent for mathematical process evaluation through {Tree-of-Error}, named StepMathAgent, which precisely evaluates all problem types while providing high interpretability and extensive feedback. 
Experimental results on our constructed StepMathBench demonstrate that StepMathAgent delivers scores highly aligned with human preferences, showcasing its adaptability across diverse scenarios.

\section*{Limitations}

This paper still has several limitations: (1) Although StepMathBench represents a highly innovative dataset for evaluating mathematical processes, its scale remains relatively small due to the complexity of its construction. This dataset needs to be further expanded, and more similar process evaluation datasets need to be constructed to support future research. (2) The external modules in StepMathAgent are specifically designed to accommodate various real-world scenarios. Nevertheless, this paper incorporates only four external modules, leaving room for the exploration of additional modules like step length recommendation and broader real-world applications. (3) Due to cost constraints, only five LLMs are selected as baselines for experimental comparisons. It is crucial to conduct a more comprehensive comparison in the future.

\bibliographystyle{plainnat}
\bibliography{main}

\begin{thebibliography}{31}
\providecommand{\natexlab}[1]{#1}
\providecommand{\url}[1]{\texttt{#1}}
\expandafter\ifx\csname urlstyle\endcsname\relax
  \providecommand{\doi}[1]{doi: #1}\else
  \providecommand{\doi}{doi: \begingroup \urlstyle{rm}\Url}\fi

\bibitem[Ahn et~al.(2024)Ahn, Verma, Lou, Liu, Zhang, and Yin]{survey-math}
Janice Ahn, Rishu Verma, Renze Lou, Di~Liu, Rui Zhang, and Wenpeng Yin.
\newblock Large language models for mathematical reasoning: Progresses and challenges, 2024.
\newblock URL \url{https://arxiv.org/abs/2402.00157}.

\bibitem[Amini et~al.(2019)Amini, Gabriel, Lin, Koncel-Kedziorski, Choi, and Hajishirzi]{MathQA}
Aida Amini, Saadia Gabriel, Shanchuan Lin, Rik Koncel-Kedziorski, Yejin Choi, and Hannaneh Hajishirzi.
\newblock {M}ath{QA}: Towards interpretable math word problem solving with operation-based formalisms.
\newblock In Jill Burstein, Christy Doran, and Thamar Solorio, editors, \emph{Proceedings of the 2019 Conference of the North {A}merican Chapter of the Association for Computational Linguistics: Human Language Technologies, Volume 1 (Long and Short Papers)}, pages 2357--2367, Minneapolis, Minnesota, June 2019. Association for Computational Linguistics.
\newblock \doi{10.18653/v1/N19-1245}.
\newblock URL \url{https://aclanthology.org/N19-1245/}.

\bibitem[Anthropic(2024)]{claude}
Anthropic.
\newblock The claude 3 model family: Opus, sonnet, haiku.
\newblock 2024.
\newblock URL \url{https://api.semanticscholar.org/CorpusID:268232499}.

\bibitem[Bai et~al.(2023)Bai, Bai, Chu, Cui, Dang, Deng, Fan, Ge, Han, Huang, and et~al.]{qwen}
Jinze Bai, Shuai Bai, Yunfei Chu, Zeyu Cui, Kai Dang, Xiaodong Deng, Yang Fan, Wenbin Ge, Yu~Han, Fei Huang, and et~al.
\newblock Qwen technical report.
\newblock \emph{arXiv preprint arXiv:2309.16609}, 2023.

\bibitem[Caraeni et~al.(2024)Caraeni, Scarlatos, and Lan]{dataset2}
Adriana Caraeni, Alexander Scarlatos, and Andrew Lan.
\newblock Evaluating gpt-4 at grading handwritten solutions in math exams, 2024.
\newblock URL \url{https://arxiv.org/abs/2411.05231}.

\bibitem[Chang et~al.(2024)Chang, Wang, Wang, Wu, Yang, Zhu, Chen, Yi, Wang, Wang, et~al.]{survey-evaluation}
Yupeng Chang, Xu~Wang, Jindong Wang, Yuan Wu, Linyi Yang, Kaijie Zhu, Hao Chen, Xiaoyuan Yi, Cunxiang Wang, Yidong Wang, et~al.
\newblock A survey on evaluation of large language models.
\newblock \emph{ACM Transactions on Intelligent Systems and Technology}, 15\penalty0 (3):\penalty0 1--45, 2024.

\bibitem[Chernyshev et~al.(2025)Chernyshev, Polshkov, Artemova, Myasnikov, Stepanov, Miasnikov, and Tilga]{dataste1}
Konstantin Chernyshev, Vitaliy Polshkov, Ekaterina Artemova, Alex Myasnikov, Vlad Stepanov, Alexei Miasnikov, and Sergei Tilga.
\newblock U-math: A university-level benchmark for evaluating mathematical skills in llms, 2025.
\newblock URL \url{https://arxiv.org/abs/2412.03205}.

\bibitem[Cobbe et~al.(2021)Cobbe, Kosaraju, Bavarian, Chen, Jun, Kaiser, Plappert, Tworek, Hilton, Nakano, Hesse, and Schulman]{GSM8K}
Karl Cobbe, Vineet Kosaraju, Mohammad Bavarian, Mark Chen, Heewoo Jun, Lukasz Kaiser, Matthias Plappert, Jerry Tworek, Jacob Hilton, Reiichiro Nakano, Christopher Hesse, and John Schulman.
\newblock Training verifiers to solve math word problems, 2021.
\newblock URL \url{https://arxiv.org/abs/2110.14168}.

\bibitem[GLM et~al.(2024)GLM, Zeng, Xu, Wang, Zhang, Yin, Rojas, Feng, Zhao, Lai, and et~al.]{glm}
Team GLM, Aohan Zeng, Bin Xu, Bowen Wang, Chenhui Zhang, Da~Yin, Diego Rojas, Guanyu Feng, Hanlin Zhao, Hanyu Lai, and et~al.
\newblock Chatglm: A family of large language models from glm-130b to glm-4 all tools, 2024.

\bibitem[Golovneva et~al.(2023)Golovneva, Chen, Poff, Corredor, Zettlemoyer, Fazel-Zarandi, and Celikyilmaz]{process5}
Olga Golovneva, Moya Chen, Spencer Poff, Martin Corredor, Luke Zettlemoyer, Maryam Fazel-Zarandi, and Asli Celikyilmaz.
\newblock Roscoe: A suite of metrics for scoring step-by-step reasoning, 2023.
\newblock URL \url{https://arxiv.org/abs/2212.07919}.

\bibitem[Grattafiori et~al.(2024)Grattafiori, Dubey, Jauhri, Pandey, Kadian, Al-Dahle, Letman, Mathur, Schelten, Vaughan, and et~al.]{llama}
Aaron Grattafiori, Abhimanyu Dubey, Abhinav Jauhri, Abhinav Pandey, Abhishek Kadian, Ahmad Al-Dahle, Aiesha Letman, Akhil Mathur, Alan Schelten, Alex Vaughan, and et~al.
\newblock The llama 3 herd of models, 2024.
\newblock URL \url{https://arxiv.org/abs/2407.21783}.

\bibitem[Hendrycks et~al.(2021)Hendrycks, Burns, Kadavath, Arora, Basart, Tang, Song, and Steinhardt]{MATH}
Dan Hendrycks, Collin Burns, Saurav Kadavath, Akul Arora, Steven Basart, Eric Tang, Dawn Song, and Jacob Steinhardt.
\newblock Measuring mathematical problem solving with the math dataset, 2021.
\newblock URL \url{https://arxiv.org/abs/2103.03874}.

\bibitem[Hu et~al.(2025)Hu, Ouyang, and Liu]{process6}
Yulan Hu, Sheng Ouyang, and Yong Liu.
\newblock Coarse-to-fine process reward modeling for enhanced mathematical reasoning, 2025.
\newblock URL \url{https://arxiv.org/abs/2501.13622}.

\bibitem[Li et~al.(2024)Li, Zhang, Yin, Ji, Bai, Pan, Zeng, Xu, Zhang, and Liu]{dataset4}
Zhong-Zhi Li, Ming-Liang Zhang, Fei Yin, Zhi-Long Ji, Jin-Feng Bai, Zhen-Ru Pan, Fan-Hu Zeng, Jian Xu, Jia-Xin Zhang, and Cheng-Lin Liu.
\newblock Cmmath: A chinese multi-modal math skill evaluation benchmark for foundation models, 2024.
\newblock URL \url{https://arxiv.org/abs/2407.12023}.

\bibitem[Lightman et~al.(2023)Lightman, Kosaraju, Burda, Edwards, Baker, Lee, Leike, Schulman, Sutskever, and Cobbe]{process1}
Hunter Lightman, Vineet Kosaraju, Yura Burda, Harri Edwards, Bowen Baker, Teddy Lee, Jan Leike, John Schulman, Ilya Sutskever, and Karl Cobbe.
\newblock Let's verify step by step, 2023.
\newblock URL \url{https://arxiv.org/abs/2305.20050}.

\bibitem[Lu et~al.(2023)Lu, Qiu, Chang, Wu, Zhu, Rajpurohit, Clark, and Kalyan]{TabMWP}
Pan Lu, Liang Qiu, Kai-Wei Chang, Ying~Nian Wu, Song-Chun Zhu, Tanmay Rajpurohit, Peter Clark, and Ashwin Kalyan.
\newblock Dynamic prompt learning via policy gradient for semi-structured mathematical reasoning, 2023.
\newblock URL \url{https://arxiv.org/abs/2209.14610}.

\bibitem[Miner et~al.(2024)Miner, Takashima, Han, Erata, Antonopoulos, Piskac, and Shapiro]{evaluation2}
Stephen Miner, Yoshiki Takashima, Simeng Han, Ferhat Erata, Timos Antonopoulos, Ruzica Piskac, and Scott~J Shapiro.
\newblock Scheherazade: Evaluating chain-of-thought math reasoning in llms with chain-of-problems, 2024.
\newblock URL \url{https://arxiv.org/abs/2410.00151}.

\bibitem[OpenAI et~al.(2024)OpenAI, Achiam, Adler, Agarwal, Ahmad, Akkaya, Aleman, Almeida, Altenschmidt, Altman, and et~al.]{gpt4}
OpenAI, Josh Achiam, Steven Adler, Sandhini Agarwal, Lama Ahmad, Ilge Akkaya, Florencia~Leoni Aleman, Diogo Almeida, Janko Altenschmidt, Sam Altman, and et~al.
\newblock Gpt-4 technical report, 2024.
\newblock URL \url{https://arxiv.org/abs/2303.08774}.

\bibitem[Song et~al.(2025)Song, Su, Qu, Zhou, and Cheng]{process8}
Mingyang Song, Zhaochen Su, Xiaoye Qu, Jiawei Zhou, and Yu~Cheng.
\newblock Prmbench: A fine-grained and challenging benchmark for process-level reward models, 2025.
\newblock URL \url{https://arxiv.org/abs/2501.03124}.

\bibitem[Sun et~al.(2025)Sun, Wang, Yu, Zang, Zheng, Xu, Zhang, Yang, and Li]{process7}
Zhongxiang Sun, Qipeng Wang, Weijie Yu, Xiaoxue Zang, Kai Zheng, Jun Xu, Xiao Zhang, Song Yang, and Han Li.
\newblock Rearter: Retrieval-augmented reasoning with trustworthy process rewarding, 2025.
\newblock URL \url{https://arxiv.org/abs/2501.07861}.

\bibitem[Team et~al.(2024)Team, Georgiev, Lei, Burnell, Bai, Gulati, Tanzer, Vincent, Pan, Wang, Mariooryad, and et~al.]{gemini}
Gemini Team, Petko Georgiev, Ving~Ian Lei, Ryan Burnell, Libin Bai, Anmol Gulati, Garrett Tanzer, Damien Vincent, Zhufeng Pan, Shibo Wang, Soroosh Mariooryad, and et~al.
\newblock Gemini 1.5: Unlocking multimodal understanding across millions of tokens of context, 2024.
\newblock URL \url{https://arxiv.org/abs/2403.05530}.

\bibitem[Uesato et~al.(2022)Uesato, Kushman, Kumar, Song, Siegel, Wang, Creswell, Irving, and Higgins]{process2}
Jonathan Uesato, Nate Kushman, Ramana Kumar, Francis Song, Noah Siegel, Lisa Wang, Antonia Creswell, Geoffrey Irving, and Irina Higgins.
\newblock Solving math word problems with process- and outcome-based feedback, 2022.
\newblock URL \url{https://arxiv.org/abs/2211.14275}.

\bibitem[Wang et~al.(2024)Wang, Li, Shao, Xu, Dai, Li, Chen, Wu, and Sui]{process3}
Peiyi Wang, Lei Li, Zhihong Shao, Runxin Xu, Damai Dai, Yifei Li, Deli Chen, Yu~Wu, and Zhifang Sui.
\newblock Math-shepherd: Verify and reinforce {LLM}s step-by-step without human annotations.
\newblock In Lun-Wei Ku, Andre Martins, and Vivek Srikumar, editors, \emph{Proceedings of the 62nd Annual Meeting of the Association for Computational Linguistics (Volume 1: Long Papers)}, pages 9426--9439, Bangkok, Thailand, August 2024. Association for Computational Linguistics.
\newblock \doi{10.18653/v1/2024.acl-long.510}.
\newblock URL \url{https://aclanthology.org/2024.acl-long.510/}.

\bibitem[Xia et~al.(2025)Xia, Li, Liu, Wu, and Liu]{process4}
Shijie Xia, Xuefeng Li, Yixin Liu, Tongshuang Wu, and Pengfei Liu.
\newblock Evaluating mathematical reasoning beyond accuracy, 2025.
\newblock URL \url{https://arxiv.org/abs/2404.05692}.

\bibitem[Yang et~al.(2025)Yang, Yang, Ma, and Liu]{evaluation1}
Bo~Yang, Qingping Yang, Yingwei Ma, and Runtao Liu.
\newblock Utmath: Math evaluation with unit test via reasoning-to-coding thoughts, 2025.
\newblock URL \url{https://arxiv.org/abs/2411.07240}.

\bibitem[Yuan et~al.(2023)Yuan, Yuan, Tan, Wang, and Huang]{MATH-140}
Zheng Yuan, Hongyi Yuan, Chuanqi Tan, Wei Wang, and Songfang Huang.
\newblock How well do large language models perform in arithmetic tasks?, 2023.
\newblock URL \url{https://arxiv.org/abs/2304.02015}.

\bibitem[Zhang et~al.(2024)Zhang, Li, and Fan]{evaluation4}
Boning Zhang, Chengxi Li, and Kai Fan.
\newblock Mario eval: Evaluate your math llm with your math llm--a mathematical dataset evaluation toolkit, 2024.
\newblock URL \url{https://arxiv.org/abs/2404.13925}.

\bibitem[Zhao et~al.(2023)Zhao, Zhou, Li, Tang, Wang, Hou, Min, Zhang, Zhang, Dong, et~al.]{survey-llm}
Wayne~Xin Zhao, Kun Zhou, Junyi Li, Tianyi Tang, Xiaolei Wang, Yupeng Hou, Yingqian Min, Beichen Zhang, Junjie Zhang, Zican Dong, et~al.
\newblock A survey of large language models.
\newblock \emph{arXiv preprint arXiv:2303.18223}, 2023.

\bibitem[Zheng et~al.(2024)Zheng, Zhang, Zhang, Lin, Lu, Yu, Liu, Zhou, and Lin]{process9}
Chujie Zheng, Zhenru Zhang, Beichen Zhang, Runji Lin, Keming Lu, Bowen Yu, Dayiheng Liu, Jingren Zhou, and Junyang Lin.
\newblock Processbench: Identifying process errors in mathematical reasoning, 2024.
\newblock URL \url{https://arxiv.org/abs/2412.06559}.

\bibitem[Zhou et~al.(2024)Zhou, Liu, Ning, Liu, Wang, Wong, Huang, Wang, and Huang]{evaluation3}
Zihao Zhou, Shudong Liu, Maizhen Ning, Wei Liu, Jindong Wang, Derek~F. Wong, Xiaowei Huang, Qiufeng Wang, and Kaizhu Huang.
\newblock Is your model really a good math reasoner? evaluating mathematical reasoning with checklist, 2024.
\newblock URL \url{https://arxiv.org/abs/2407.08733}.

\bibitem[Zou et~al.(2024)Zou, Wang, Thakur, and Kani]{dataset3}
Jiaru Zou, Qing Wang, Pratyush Thakur, and Nickvash Kani.
\newblock Stem-pom: Evaluating language models math-symbol reasoning in document parsing, 2024.
\newblock URL \url{https://arxiv.org/abs/2411.00387}.

\end{thebibliography}

\appendix

\section{Dataset and Prompts}
\label{sec:appendix dataset}

Our constructed StepMathBench is a novel Chinese mathematical process benchmark consists of 1,000 step-divided process evaluation instances and 200 high-quality math problems, which is divided by problem type, subject category and difficulty level. 
The complete dataset and code are available at https://github.com/SHU-XUN/StepMathAgent. The prompts utilized in the experiments are provided in Table \ref{tab5: prompt}, where \textit{Solution Generation} means the prompts used for generating solutions to 200 problems by five LLMs, \textit{Baseline} means the implementation prompts for the three baseline versions of LLM-based scoring. Meanwhile, \textit{\textbf{StepMathAgent}} means the implementations of its four internal operations as well as the implementations that incorporate each of the four external modules.

\begin{CJK}{UTF8}{gbsn}
\begin{xltabular}{\textwidth}{|l|l|X|}
         \hline \textbf{Purpose} & \textbf{Version} & \textbf{Prompt} \\ \hline 
        \endfirsthead
        
        \hline \multicolumn{3}{|r|}%
        {Continued from previous page} \\
        \hline \textbf{Purpose} & \textbf{Version} & \textbf{Prompt} \\ \hline 
        \endhead
        
        \hline \multicolumn{3}{|r|}{{Continued on next page}} \\ \hline
        \endfoot
        
        \hline
        \endlastfoot

         \multirow{2}{*}[-8ex]{\parbox{1.5cm}{Solution \\Generation}} & Calculation & 你是一名数学领域的专家，请严格按照如下格式回答问题：“解题过程：\textbackslash n【XXX】\textbackslash n\textbackslash n最终答案：\textbackslash n【YYY】”。其中，YYY为你的最终答案，请用一个【】符号将最终答案框起来。最终答案的具体形式需要遵从题目中的答案限定条件。例如，解题过程：\textbackslash n【a=2+1=3, b=a-2=1】\textbackslash n\textbackslash n最终答案：\textbackslash n【3,1】。\textbackslash n\textbackslash n下面请开始回答问题。\\
         \cmidrule(){2-3}
         & \parbox{1.8cm}{Proof or \\ Open-ended} & 你是一名数学领域的专家，请回答如下数学问题。\\
         \midrule
         \multirow{3}{*}[-30ex]{Baseline} & V1 & 你是一名专业的数学评分专家，擅长客观评价数学题目的回复质量。数学题目共有三种类型，分别是计算题、证明题、开放题。现在，给你提供一个数学问题和一个参考答案，请基于参考答案判断回复内容中的最终答案是否正确。若回复内容的答案与参考答案一致，则认为正确给1分，否则认为错误给0分。请注意，只需判断回复内容的结果是否正确，无需关注解题过程的正确与否。证明题无参考答案，请自行判断回复内容是否正确；开放题的答案不一，参考答案中只给出了一种情况，其他情况也请自行判断。在分析完毕后，请另起一行，返回一个标准json格式的答案，即：\{\textbackslash "score\textbackslash ": 0/1\}。\\
         \cmidrule(){2-3}
         & V2 & 你是一名专业的数学评分专家，擅长按照解题过程客观地评价数学题目的回复质量。数学题目共有三种类型，分别是计算题、证明题、开放题。现在，给你提供一个数学问题和一个参考答案，请逐步分析回复内容中的解题步骤和最终答案，并对其进行综合打分。请注意，打分区间为0-10分，且证明题无参考答案，开放题参考答案不一。在分析完毕后，请另起一行，返回一个标准json格式的答案，即：\{\textbackslash "score\textbackslash ": 5\}。 \\
         \cmidrule(){2-3}
         & V3 & 你是一名专业的数学评分专家，擅长按照解题过程客观地评价数学题目的回复质量。数学题目共有三种类型，分别是计算题、证明题、开放题。现在，给你提供一个数学问题和一个参考答案，请逐步分析回复内容中的解题步骤和最终答案，并对其进行综合打分。在打分过程中，可以从如下几个维度进行评测：答案正确性、过程正确性、方案合理性、表述清晰性、指令遵循性、整体完备性，最终给出一个综合得分。请注意，打分区间为0-10分，且证明题无参考答案，开放题参考答案不一。在分析完毕后，请另起一行，返回一个标准json格式的答案，即：\{\textbackslash "score\textbackslash ": 5\}。\\
         \multirow{6}{*}[-52ex]{\parbox{1.5cm}{\textbf{StepMath} \\ \textbf{-Agent}}} & / & {你是一名专业的数学评分专家，擅长按照解题过程客观地评价数学题目的回复质量。数学题目共有三种类型，分别是计算题、证明题、开放题。现在，给你提供一个数学问题和一个参考答案，请首先将回复内容按照推理步骤进行划分，并确保划分出的每个推理步骤都是最细粒度的，如果是计算题的话最终答案一般为划分步骤中的最后一个推理步骤。然后，请依次判断每一个划分出的推理步骤是否正确，正确则为1，错误则为0。紧接着，根据如下的计算公式计算出这个回复的最终得分，假设划分出的推理步骤共有n步，则计算题的最终得分S为：S=6*(前n-1步中正确的推理步骤)/(n-1)+4*第n个推理步骤得分，证明题和开放题的最终得分S为：S=10*(n步中正确的推理步骤)/n。最终得分需要进行四舍五入取值，仅保留整数位。最后，请输出这道题目中所有的错误链，错误链由划分出的错误的推理步骤序号组成，如(3)-(4)-(6)。请注意：1.最终得分应该在0-10分之间；2.如果中间某个步骤单独来看是正确的，但由于之前的推理步骤出错导致这个推理步骤的正确没有意义，此时这个步骤的得分为0；3.错误链应包含没有意义的推理步骤，且应列举所有的错误链使其可以构成错误树；4.请在分析完毕之后，另起一行，返回一个标准json格式的答案，如：\{\textbackslash "（1）具体的推理步骤1...\textbackslash ": 1, \textbackslash "（2）具体的推理步骤2...\textbackslash ": 0, ..., \textbackslash "（n）具体的推理步骤n...\textbackslash ": 0, \textbackslash "最终得分\textbackslash ": 7, \textbackslash "错误链\textbackslash ": \textbackslash "(3)-(4)-(6), (5)-(6)\textbackslash "\}。现在，请开始。} \\
         \cmidrule(){2-3}
         & + Difficulty & 你是一名专业的数学评分专家，擅长按照解题过程客观地评价数学题目的回复质量。数学题目共有三种类型，分别是计算题、证明题、开放题。现在，给你提供一个数学问题和一个参考答案，请首先将回复内容按照推理步骤进行划分，并确保划分出的每个推理步骤都是最细粒度的，如果是计算题的话最终答案一般为划分步骤中的最后一个推理步骤。然后，请依次判断每一个划分出的推理步骤是否正确，正确则为1，错误则为0。紧接着，根据如下的计算公式计算出这个回复的最终得分，假设划分出的推理步骤共有n步，则计算题的最终得分S为：S=6*(前n-1步中正确的推理步骤)/(n-1)+4*第n个推理步骤得分，证明题和开放题的最终得分S为：S=10*(n步中正确的推理步骤)/n。最终得分需要进行四舍五入取值，仅保留整数位。最后，请输出这道题目中所有的错误链，错误链由划分出的错误的推理步骤序号组成，如(3)-(4)-(6)。请注意：1.最终得分应该在0-10分之间；2.如果中间某个步骤单独来看是正确的，但由于之前的推理步骤出错导致这个推理步骤的正确没有意义，此时这个步骤的得分为0；3.错误链应包含没有意义的推理步骤，且应列举所有的错误链使其可以构成错误树；4.请在分析完毕之后，另起一行，返回一个标准json格式的答案，如：\{\textbackslash "（1）具体的推理步骤1...\textbackslash ": 1, \textbackslash "（2）具体的推理步骤2...\textbackslash ": 0, ..., \textbackslash "（n）具体的推理步骤n...\textbackslash ": 0, \textbackslash "最终得分\textbackslash ": 7, \textbackslash "错误链\textbackslash ": \textbackslash "(3)-(4)-(6), (5)-(6)\textbackslash "\}；5.特别地，如果某道题目特别简单而无需进行过程评估，可仅判断答案是否正确并输出0或10的最终得分和错误链，即：\{\textbackslash "最终得分\textbackslash ": 0/10, \textbackslash "错误链\textbackslash ": \textbackslash "\textbackslash "\},至于题目是否足够简单到无需进行过程评估，请自行判断。现在，请开始。\\
         & + Simplicity & 你是一名专业的数学评分专家，擅长按照解题过程客观地评价数学题目的回复质量。数学题目共有三种类型，分别是计算题、证明题、开放题。现在，给你提供一个数学问题和一个参考答案，请首先将回复内容按照推理步骤进行划分，并确保划分出的每个推理步骤都是最细粒度的，如果是计算题的话最终答案一般为划分步骤中的最后一个推理步骤。然后，请依次判断每一个划分出的推理步骤是否正确，正确则为1，错误则为0。紧接着，根据如下的计算公式计算出这个回复的最终得分，假设划分出的推理步骤共有n步，则计算题的最终得分S为：S=6*(前n-1步中正确的推理步骤)/(n-1)+4*第n个推理步骤得分，证明题和开放题的最终得分S为：S=10*(n步中正确的推理步骤)/n。最终得分需要进行四舍五入取值，仅保留整数位。最后，请输出这道题目中所有的错误链，错误链由划分出的错误的推理步骤序号组成，如(3)-(4)-(6)。请注意：1.最终得分应该在0-10分之间；2.如果中间某个步骤单独来看是正确的，但由于之前的推理步骤出错导致这个推理步骤的正确没有意义，此时这个步骤的得分为0；3.错误链应包含没有意义的推理步骤，且应列举所有的错误链使其可以构成错误树；4.特别注意的是，在依次判断划分出的推理步骤是否正确时，如果某个步骤正确但属于没有实际意义的废话或过于累赘，请判断该步骤错误；5.请在分析完毕之后，另起一行，返回一个标准json格式的答案，如：\{\textbackslash "（1）具体的推理步骤1...\textbackslash ": 1, \textbackslash "（2）具体的推理步骤2...\textbackslash ": 0, ..., \textbackslash "（n）具体的推理步骤n...\textbackslash ": 0, \textbackslash "最终得分\textbackslash ": 7, \textbackslash "错误链\textbackslash ": \textbackslash "(3)-(4)-(6), (5)-(6)\textbackslash "\}。现在，请开始。\\
         \cmidrule(){2-3}
         & + Completeness & 你是一名专业的数学评分专家，擅长按照解题过程客观地评价数学题目的回复质量。数学题目共有三种类型，分别是计算题、证明题、开放题。现在，给你提供一个数学问题和一个参考答案，请首先将回复内容按照推理步骤进行划分，并确保划分出的每个推理步骤都是最细粒度的，如果是计算题的话最终答案一般为划分步骤中的最后一个推理步骤。然后，请依次判断每一个划分出的推理步骤是否正确，正确则为1，错误则为0。紧接着，根据如下的计算公式计算出这个回复的最终得分，假设划分出的推理步骤共有n步，则计算题的最终得分S为：S=6*(前n-1步中正确的推理步骤)/(n-1)+4*第n个推理步骤得分，证明题和开放题的最终得分S为：S=10*(n步中正确的推理步骤)/n。最终得分需要进行四舍五入取值，仅保留整数位。最后，请输出这道题目中所有的错误链，错误链由划分出的错误的推理步骤序号组成，如(3)-(4)-(6)。请注意：1.最终得分应该在0-10分之间；2.如果中间某个步骤单独来看是正确的，但由于之前的推理步骤出错导致这个推理步骤的正确没有意义，此时这个步骤的得分为0；3.错误链应包含没有意义的推理步骤，且应列举所有的错误链使其可以构成错误树；4.特别注意的是，在依次判断划分出的推理步骤是否正确时，请同时考虑推理步骤的完整性和正确性并进行严格打分，只有当前提和结论都存在且推理过程正确严谨的时候才能判断正确；5.请在分析完毕之后，另起一行，返回一个标准json格式的答案，如：\{\textbackslash "（1）具体的推理步骤1...\textbackslash ": 1, \textbackslash "（2）具体的推理步骤2...\textbackslash ": 0, ..., \textbackslash "（n）具体的推理步骤n...\textbackslash ": 0, \textbackslash "最终得分\textbackslash ": 7, \textbackslash "错误链\textbackslash ": \textbackslash "(3)-(4)-(6), (5)-(6)\textbackslash "\}。现在，请开始。\\
         & + Format & 你是一名专业的数学评分专家，擅长按照解题过程客观地评价数学题目的回复质量。数学题目共有三种类型，分别是计算题、证明题、开放题。现在，给你提供一个数学问题和一个参考答案，请首先将回复内容按照推理步骤进行划分，并确保划分出的每个推理步骤都是最细粒度的，如果是计算题的话最终答案一般为划分步骤中的最后一个推理步骤。然后，请依次判断每一个划分出的推理步骤是否正确，正确则为1，错误则为0。紧接着，根据如下的计算公式计算出这个回复的最终得分，假设划分出的推理步骤共有n步，则计算题的最终得分S为：S=6*(前n-1步中正确的推理步骤)/(n-1)+4*第n个推理步骤得分，证明题和开放题的最终得分S为：S=10*(n步中正确的推理步骤)/n。最终得分需要进行四舍五入取值，仅保留整数位。最后，请输出这道题目中所有的错误链，错误链由划分出的错误的推理步骤序号组成，如(3)-(4)-(6)。请注意：1.最终得分应该在0-10分之间；2.如果中间某个步骤单独来看是正确的，但由于之前的推理步骤出错导致这个推理步骤的正确没有意义，此时这个步骤的得分为0；3.错误链应包含没有意义的推理步骤，且应列举所有的错误链使其可以构成错误树；4.特别注意的是，在依次判断划分出的推理步骤是否正确时，如果推理步骤中存在如latex或其他形式的公式，请同时考虑推理步骤的逻辑正确性和格式正确性并进行严格打分，只有逻辑和格式都正确的时候才能判断该步骤正确；5.请在分析完毕之后，另起一行，返回一个标准json格式的答案，如：\{\textbackslash "（1）具体的推理步骤1...\textbackslash ": 1, \textbackslash "（2）具体的推理步骤2...\textbackslash ": 0, ..., \textbackslash "（n）具体的推理步骤n...\textbackslash ": 0, \textbackslash "最终得分\textbackslash ": 7, \textbackslash "错误链\textbackslash ": \textbackslash "(3)-(4)-(6), (5)-(6)\textbackslash "\}。现在，请开始。\\
         & + All & 你是一名专业的数学评分专家，擅长按照解题过程客观地评价数学题目的回复质量。数学题目共有三种类型，分别是计算题、证明题、开放题。现在，给你提供一个数学问题和一个参考答案，请首先将回复内容按照推理步骤进行划分，并确保划分出的每个推理步骤都是最细粒度的，如果是计算题的话最终答案一般为划分步骤中的最后一个推理步骤。然后，请依次判断每一个划分出的推理步骤是否正确，正确则为1，错误则为0。紧接着，根据如下的计算公式计算出这个回复的最终得分，假设划分出的推理步骤共有n步，则计算题的最终得分S为：S=6*(前n-1步中正确的推理步骤)/(n-1)+4*第n个推理步骤得分，证明题和开放题的最终得分S为：S=10*(n步中正确的推理步骤)/n。最终得分需要进行四舍五入取值，仅保留整数位。最后，请输出这道题目中所有的错误链，错误链由划分出的错误的推理步骤序号组成，如(3)-(4)-(6)。请注意：1.最终得分应该在0-10分之间；2.如果中间某个步骤单独来看是正确的，但由于之前的推理步骤出错导致这个推理步骤的正确没有意义，此时这个步骤的得分为0；3.错误链应包含没有意义的推理步骤，且应列举所有的错误链使其可以构成错误树；4.特别注意的是，请在整个评分过程中额外考虑简洁性、完整性、格式正确性三个维度，其中：- 简洁性指的是如果某个步骤正确但属于没有实际意义的废话或过于累赘，请判断该步骤错误，- 完整性指的是在依次判断划分出的推理步骤是否正确时，请同时考虑推理步骤的完整性和正确性并进行严格打分，只有当前提和结论都存在且推理过程正确严谨的时候才能判断正确，- 格式正确性指的是在依次判断划分出的推理步骤是否正确时，如果推理步骤中存在如latex或其他形式的公式，请同时考虑推理步骤的逻辑正确性和格式正确性并进行严格打分，只有逻辑和格式都正确的时候才能判断该步骤正确；5.请在分析完毕之后，另起一行，返回一个标准json格式的答案，如：\{\textbackslash "（1）具体的推理步骤1...\textbackslash ": 1, \textbackslash "（2）具体的推理步骤2...\textbackslash ": 0, ..., \textbackslash "（n）具体的推理步骤n...\textbackslash ": 0, \textbackslash "最终得分\textbackslash ": 7, \textbackslash "错误链\textbackslash ": \textbackslash "(3)-(4)-(6), (5)-(6)\textbackslash "\}；6.特别地，如果某道题目特别简单而无需进行过程评估，可仅判断答案是否正确并输出0或10的最终得分和错误链，即：\{\textbackslash "最终得分\textbackslash ": 0/10, \textbackslash "错误链\textbackslash ": \textbackslash "\textbackslash "\},至于题目是否足够简单到无需进行过程评估，请自行判断。现在，请开始。\\
\end{xltabular}
\captionof{table}{All the prompts utilized in the experiments.}
\label{tab5: prompt}
\end{CJK}

\section{Case Study}
\label{sec: apppendix case study}

\begin{figure*}[t]
  \centering
  \includegraphics[width=\textwidth]{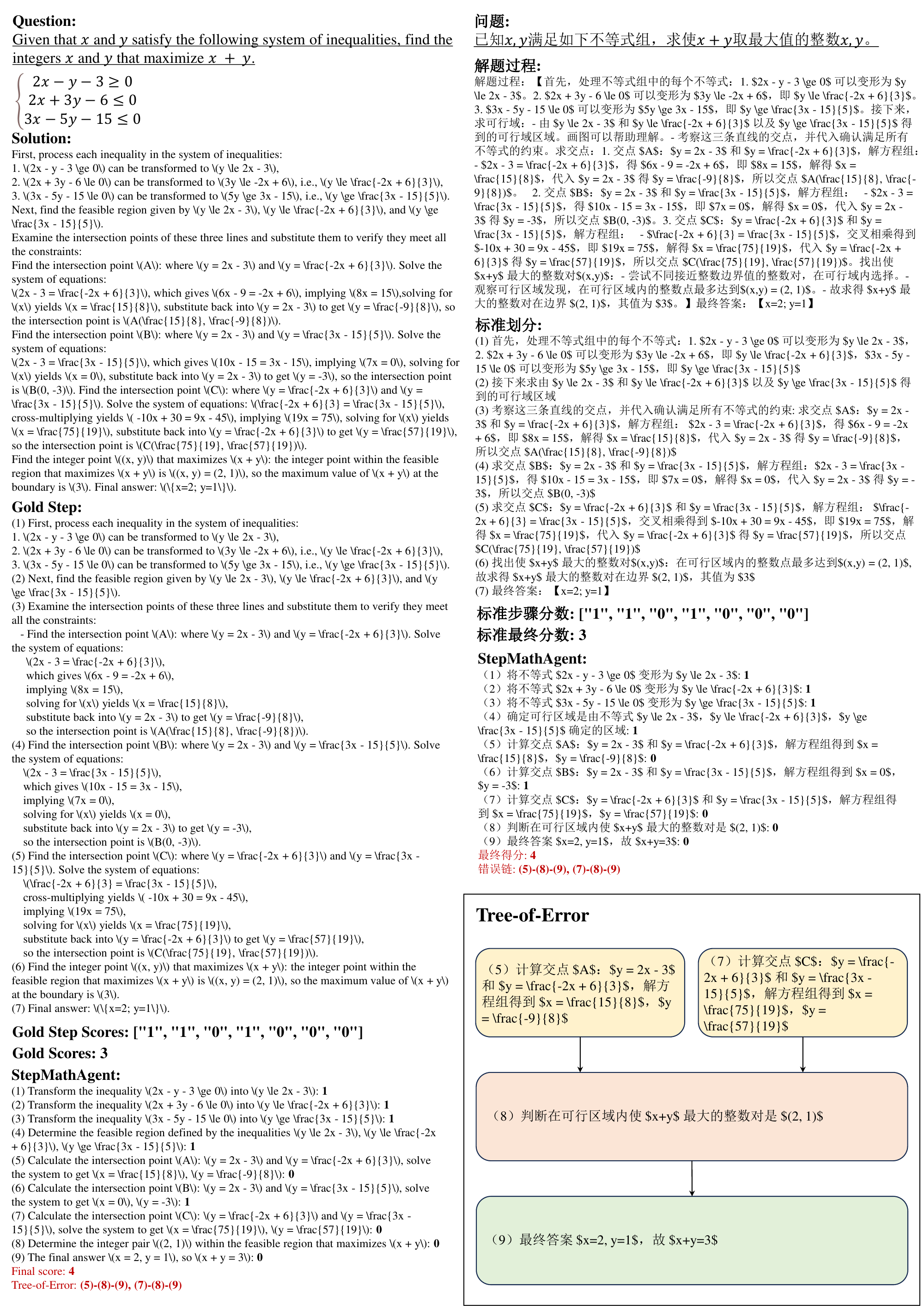}
  \caption{Case Study on StepMathAgent. The Chinese text on the right represents the original version, while the corresponding English translation is presented on the left.}
  \label{fig5-case}
\end{figure*}

We select several representative cases from StepMathBench to comprehensively evaluate the effectiveness and interpretability of the proposed StepMathAgent, as shown in Figure \ref{fig5-case}.

This case demonstrates that StepMathAgent achieves a high degree of alignment with human evaluations across all stages, including logical step segmentation, step scoring, score aggregation and error tree generation. Although the segmented steps are slightly longer, the final scores align closely with human judgments. Moreover, the scoring process and the generated error tree offer exceptional interpretability and comprehensive feedback.

\end{document}